\documentclass{article}

\usepackage{arxiv}

\usepackage[utf8]{inputenc} 
\usepackage[T1]{fontenc}    
\usepackage{hyperref}       
\usepackage{url}            
\usepackage{booktabs}       
\usepackage{amsfonts}       
\usepackage{nicefrac}       
\usepackage{microtype}      
\usepackage{lipsum}		
\usepackage{graphicx}
\usepackage{natbib}
\usepackage{doi}

\usepackage{paper_packages}

\title{A Review of Open Source Software Tools for Time Series Analysis}

\author{ \href{https://orcid.org/0000-0002-3584-8533}{\includegraphics[scale=0.06]{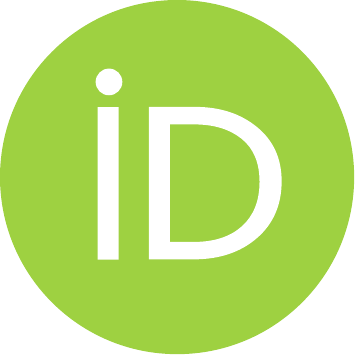}\hspace{1mm}Yunus Parvej Faniband} \\
	  Office of Industrial Collaboration\\
	 King Fahd University of Petroleum \& Minerals\\
	Dhahran, Saudi Arabia 312613 \\
	\texttt{yparvej@kfupm.edu.sa} \\
   \And
   \href{https://orcid.org/0000-0001-8874-1417}{\includegraphics[scale=0.06]{orcid.pdf}\hspace{1mm}Iskandar Ishak} \\
 		Faculty of Computer Science and Information Technology\\
	 	Universiti Putra Malaysia\\
		43400 UPM, Serdang, Selangor Darul Ehsan, Malaysia \\
	 \texttt{iskandar\_i@upm.edu.my} \\ 
	\AND
	\href{https://orcid.org/0000-0002-4796-0581}{\includegraphics[scale=0.06]{orcid.pdf}\hspace{1mm}Sadiq Sait} \\
	Office of Industrial Collaboration\\
	King Fahd University of Petroleum \& Minerals\\
	Dhahran, Saudi Arabia 312613 \\
	\texttt{sadiq@kfupm.edu.sa} \\
}

\renewcommand{\shorttitle}{\textit{arXiv} Template}

\hypersetup{
pdftitle={A template for the arxiv style},
pdfsubject={q-bio.NC, q-bio.QM},
pdfauthor={David S.~Hippocampus, Elias D.~Striatum},
pdfkeywords={First keyword, Second keyword, More},
}

\begin{document}
\maketitle
\begin{abstract}
	
	Time series data is used in a wide range of real-world applications. In a variety of domains, detailed analysis of time series data (via Forecasting and Anomaly Detection) leads to a better understanding of how events associated with a specific time instance behave. Time Series Analysis (TSA) is commonly performed with plots and traditional models. Machine Learning (ML) approaches, on the other hand, have seen an increase in the state of the art for Forecasting and Anomaly Detection because they provide comparable results when time and data constraints are met. A number of time series toolboxes are available that offer rich interfaces to specific model classes (ARIMA/filters, neural networks) or framework interfaces to isolated time series modelling tasks (forecasting, feature extraction, annotation, classification). Nonetheless, open-source machine learning capabilities for time series remain limited, and existing libraries are frequently incompatible with one another. The goal of this paper is to provide a concise and user-friendly overview of the most important open-source tools for time series analysis. This article examines two related toolboxes: (1) forecasting and (2) anomaly detection. This paper describes a typical Time Series Analysis (TSA) framework with an architecture and lists the TSA framework’s main features. The tools are categorized based on the criteria of analysis tasks completed, data preparation methods employed, and evaluation methods for results generated. This paper presents quantitative analysis and discusses the current state of actively developed open-source Time Series Analysis frameworks. Overall, this article considered 60 time series analysis tools, and 32 of which provided forecasting modules, and 21 packages included anomaly detection. 

\end{abstract}

\keywords{Time Series Analysis \and Forecasting  \and Anomaly detection \and Open Source Software \and}

\section{Introduction}

Time series are defined as “a collection of observations made sequentially through time”  \cite{chatfield2005time}.  An important area of statistical analysis is the methods of time series analysis. The study of events over time occurs in a wide range of fields ranging from marketing to finance and engineering, but also for IoT data, climate reading and renewable energy \cite{faniband2021univariate} \cite{faniband2020forecasting} . Identifying how specific aspects perform, and being ready to foresee their expansion, aids to diagnose issues, foresee actions, and conclusively strengthen greater outcomes. Although this leads to several specifications, forecast of instances based on prior behavior (forecasting) and the detection of sporadic observations, in relationship to the generalization of the data items (anomaly detection) are two familiar areas of time series analysis.

Categorization of time series becomes necessary because no approach has the necessary generalization capabilities to address every type of time series. Seasonality, Trend, and Outliers are the main features to consider when classifying time series \cite{chatfield2003the}. The instances of time series analysis scenarios constitutes an important problem in areas such as sociology, scientific science, economy, economics, stock control, marketing, production planning and engineering as argued by the authors in \cite{brockwell2016introduction}. The objective of the analysis of the time series is to draw conclusions from these structures. Following the establishment of a convenient model collection, parameters can be evaluated, data fitness can be investigated, and fitted models may be used to gain a better understanding of the series generating process. The models can be applied to characterize data, extract noise from signals (via time plots) and test hypotheses, such as global warming using climate data. These goals can be divided into four categories: description, explanation, prediction (forecasting), and control.

Rapid Prototyping, Reproducibility and Transparency are the three main areas where Toolboxes aid the researcher community. Toolboxes are essential for Rapid Prototyping, as they provide a quick and efficient way to implement and experiment with new models. Users and researchers can evaluate and compare models quickly and systematically. The other critical feature is Reproducibility, which allows researchers to reliably reproduce the results of existing models and then compare them with new models. Machine learning and forecasting research are very much dependent on reproducibility \cite{makridakis2018objectivity} \cite{arnold2019turing}. Toolboxes make algorithms and workflows more readable and transparent by providing a consistent interface for algorithms and composition functionality.

Forecasting is the prediction of unknown values inside an observable time series. For example, given an observed series x$_1$, x$_2$,..., x$_n$, the objective is to estimate values, x$_{n+h}$, where \emph{h} is the forecasting horizon. This is a significant difficulty in various aspects of the economy, such as inventory control, market strategy, and production planning. In real-world applications, time series forecasting is omnipresent. One example is forecasting of demand to supply inventory space, along with long-term economic growth forecasts to inform government policies. Other examples include predicting stock prices to aid financial decisions, and forecasting of demand to fill up inventory. Forecasting is also trending area for machine learning research, with recent advances in pure and hybrid machine learning approaches \cite{smyl2020hybrid}. Forecasting in practice entails a number of steps: we must first specify, fit, and select an appropriate model, before evaluating and deploying it. We can implement these steps using various open-source toolboxes. In most cases, existing toolboxes tend to be severely limited in key areas. While some models support specific model families (e.g. Autoregressive Integrated Moving Average (ARIMA) or neural networks), others support more generic frameworks for forecasting. Some other options offer functionality only for specific parts of the process (e.g. feature extraction). Others provide more generic forecasting frameworks but lack interfaces to well-known machine learning toolkits such as scikit-learn \cite{pedregosa2011scikit}. It appears, however, that despite the successful implementation of machine learning for forecasting, there is a lack of open-source toolbox that connects existing machine learning packages, enabling model development, tuning, and evaluation.

Anomaly detection is a "important analysis task, which detects anomals or abnormals in a certain set of data" \cite{ahmed2016survey}. It is also known as outlier or novelty detection and Outliers are essential because they illustrate uncommon events in a number of fields, which commonly speed up decisive actions to address. Recent trends indicate a sharp increase in the number of computer systems and digital gadgets over the last decade, due to the wide range of technologies present in the same time period. As the networks grew, security concerns increased and It is expected that unusual network traffic may be indicative of a compromised device. A common anomaly in a credit card report could indicate fraud, for instance. If there are visual anomalies in an MRI image, it could indicate the presence of a tumor. Similarly anomalies in frequencies on an EEG scan indicate seizures and/or their pre/post conditions. Anomalies from a wide variety of areas can be extrapolated and are categorized into following three groups namely Point Anomaly, Contextual Anomaly and Collective Anomaly \cite{ahmed2016surveyfindomain}.


Based on the literature, performing time series analysis can be a time-consuming endeavor \cite{petropoulos2022forecasting}. In actuality, it's not as simple as writing an algorithm and handing it over to a machine. To begin, the data must be prepared and cleaned. Manually conducting repetitive, time-consuming, and hands-on trial and error experiments must follow data processing. TSA frameworks, on the other hand, must be scalable and accommodate a variety of time series approaches, as well as automate them and allow them to be used iteratively to develop the model. The current ecosystem of Tools is fragmented , with many specialized tool kits, but no overarching framework, and difficult to understand, use and inter-operate.


Following are the contributions of this article:
\begin{itemize}
	\item Introduces the basic requirements for a typical Time Series Analysis (TSA) framework ( Section \ref{section_functional_specs_tsa_framework}) with architecture and describe the main features of TSA framework in relation to the differences expressed between Univariate Time Series (UTS) and Multivariate Time Series (MTS) approaches (Section \ref{section_tsa_framework}) .
	\item A list of Tools for TSA which provide automation in time series analysis, for forecasting and anomaly detection ( See Section \ref{section_tsa_tools})
	\item This paper presents quantitative analysis and discusses the state of the current actively developed open-source frameworks for Time Series Analysis (Section \ref{section_discussion}), including Forecasting, and Anomaly Detection
\end{itemize}

\section{Methodology}
This work conducted systematic literate review according to the guidelines specified in \cite{kitchenham2013systematic}. These guidelines, on the other hand, are geared toward printed materials (academic research articles) rather than software packages. As a result, some adjustments are made to these procedures and explored the literature databases and open source code repositories (e.g., Bitbucket, SourceForge, Github).

\begin{figure}
	\centering
	\begin{tikzpicture}  
		\small
		\node[blockrounded] (a) {Literature Search \emph{(SC:1, 3)} };  
		\node[block,right=of a] (b) {Checking Open Source Code Repository (e.g., Bitbucket, SourceForge, Github) \emph{(SC2.*})};   
		\node[block,right=of b] (c) {Removing duplicates};  
		
		\node [decision, right=of c] (d) {+};
		\node[blockrounded] (e) at ([yshift=-2cm]$(a)!1.0!(b)$) {Analysis};   
		\node[block] (f) at ([yshift=-2cm]$(b)!1.0!(c)$) {Generic vs Domain specific \emph{(SC4)}};  
		\node[block] (f1) at ([yshift=-2cm]$(c)!1.0!(d)$) {Snowballing \emph{(SC1, 2.*, 3)}};  
		\node[block] (g) at ([yshift=2cm]$(c)!1.0!(d)$) {Focuses on Time Series Analysis? \emph{(SC3)}};   
		\node[block] (h1) at ([yshift=2cm]$(b)!1.0!(b)$) {Removing duplicates};  
		\node[block] (g1) at ([yshift=2cm]$(c)!1.0!(c)$) {Is Tool implented in Python, Java, R or Julia  ? \emph{(SC2.3)} };  
		\node[blockrounded] (h) at ([yshift=2cm]$(a)!1.0!(a)$) {Open Source Code Repository (e.g.,  Bitbucket, SourceForge, Github) Search \emph{(SC:1, 2.1, 2.2)}};   
		\draw [line] (a) -- node[name=a2b,above] {$ 112 $} (b);
		\draw[line] (b)-- node[name=b2c,above] {$ 22 $}(c);  
		\draw[line] (c)-- node[name=c2d,above] {$ 12 $} (d);  
		\draw[line] (f)-- node[name=f2e,above] {$60$} (e);  
		\draw[line] (f1)-- node[name=f12f,above] {$ 79 $} (f);  
		
		\draw[line] (h)-- node[name=h2h1,above] {$ 127 $} (h1); 
		\draw[line] (h1)-- node[name=h12g1,above] {$ 90 $} (g1); 
		\draw[line] (g1)-- node[name=g12g,above] {$ 71 $} (g); 
		\draw[line] (g.south) -- node[name=g2d,right] {$ 48 $} (d); 
		
		\draw[line] (d.south) -- node[name=d2f1,left] {$ 60 $} (f1);  	
	\end{tikzpicture}  
	
	\caption{Overview of the search and filtering procedure for Tools code repository. The number of repositories left after each stage is indicated by edge labels.} 
	\label{fig:search_filtering_method}
\end{figure}
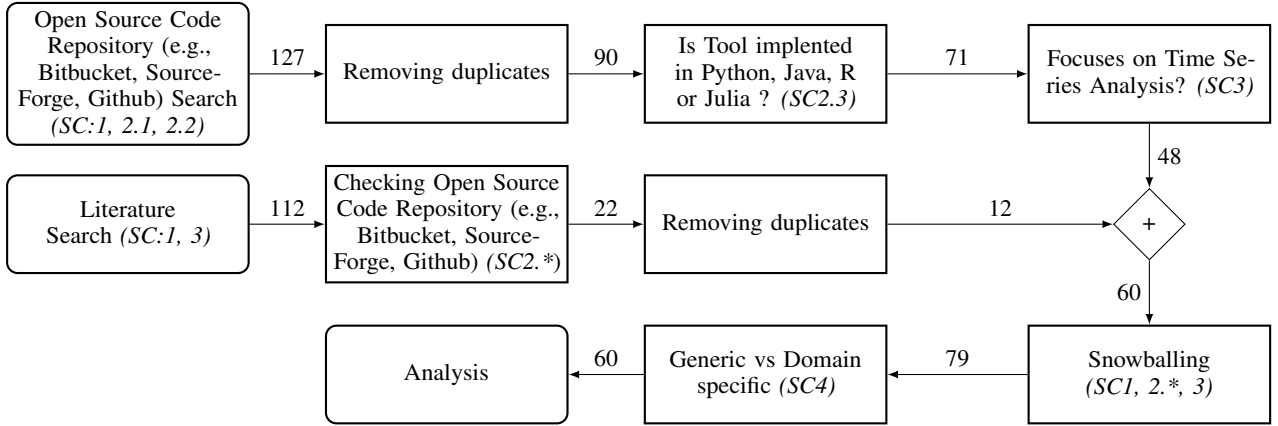	

\subsection{Selection Criteria for Tools:}

Based on the literature review the TSA tasks fall into following categories:  Forecasting (Prediction) (\emph{TSK1})   ( \cite{fakhrazari2017survey} \cite{hendikawati2020survey} \cite{mahalakshmi2016survey} \cite{panigrahi2020fuzzy} \cite{tealab2018time} ),   Classification  (\emph{TSK2}) (\cite{abanda2019review} \cite{bagnall2017great}), Clustering  (\emph{TSK3}) (\cite{aghabozorgi2015time}), Anomaly Detection (\emph{TSK4}) (\cite{wu2016survey}, \cite{cook2019anomaly}, \cite{ayadi2017outlier} , Segmentation (Summarization) (TSK5), Pattern Recognition (\emph{TSK6}) ( \cite{torkamani2017survey}, \cite{wang2019deep}) and Change Point Detection(\emph{TSK7}) ( \cite{sharma2016trend}, \cite{truong2020selective}, \cite{aminikhanghahi2017survey} ). This research work consider the open source tools implementing these tasks.

In terms of providing the data preparation capabilities, tools providing following set of methods are considered :  Dimensionality reduction methods (\emph{DP1}), Tools explicitly providing missing values imputation methods (\emph{DP2}), tools serving decomposition methods  (\emph{DP3}) (e.g., decomposing time series into trends, seasonal components, or frequency components), tools integrated with generic transformation and features generation methods  (\emph{DP4}) and tools providing methods for computing similarity measures  (\emph{DP5}).

With respect to the Evaluation criteria followed by the TSA Tools, the following categories are identified : Tools providing the methods for model selection, hyperparameter search, or feature selection (\emph{EVL1}),  tools Providing evaluation metrics and statistical tests(\emph{EVL2}) and tools Providing visualization methods (\emph{EVL3}). Some of the tools provide functions for either generating synthetic time series data (\emph{DS1}) or integrate functionality to download existing datasets (\emph{DS2}). 

The tools should be open source to explicitly target time series analysis (SC3)  and implemented using popular programming language like Python,Java, C++, Julia  or R (SC1). Some tools that can be used as building blocks for time series analysis and whose primary purpose is not time series analysis per se were excluded from consideration (e.g., scikit-learn, scipy , numpy in case of Python). The search process narrowed down to tools that provide methods which are generally domain-agnostic (IC4), and domain-specific tools were excluded.

With respect to source code repository of open source tools following points are considered. The tools should have been evolved over a time with continuous improvement and actively maintained(SC2.1) (last commit within less than 6 months). For source repositories available in Github , the tool should have more than 100 GitHub stars (SC2.2).  For python based tools, it should be listed in PyPI and be installable via pip or conda (SC2.3).

\section{Functional Specifications of TSA framework}
\label{section_functional_specs_tsa_framework}
The four baseline functional specifications of a TSA framework are as follows \cite{costa2019time}( See Appendix:Table \ref{tab:requirments}).
\begin{enumerate}
	\item \textbf{Data processing capabilities}: The framework must be able to provide and apply data cleaning and preparation strategies for the input dataset.
	\item \textbf{A Forecasting framework}: The framework must be able to select an adequately suited approach based on the characteristics of the given time series, which will cover both basic and optimized approaches to forecasting for single and multi-value time series.
	\item \textbf{Anomaly Detection framework}: The framework must be able to detect anomalous or abnormal data from a given dataset employing specialized anomaly detection approaches for single and multi-value time series.
	\item \textbf{Exposing framework operations through REST API}: The framework must include a REST API that supports all operations required for forecasting and anomaly detection.
\end{enumerate}

\paragraph*{Build capabilities for data processing (FS.01)}

Certain time series analysis techniques require the data to contain, or not contain, specific characteristics. Some time series are classified according to seasonal patterns, trends, and anomalies (cite references ). The necessity for implementing techniques that identify and, if necessary, remove seasonal and trend patterns is enforced by this requirement. This specification also incorporates the handling of other time series characteristics, including missing values and noise.

\begin{enumerate}

	\item \textbf{Physical data processing characteristics}: It is critical that the data be formatted properly for the framework operations. Keeping this in mind, this requirement is in charge of resolving inconsistencies caused by data entry errors.

	\item \textbf{Processing time series data characteristics} : To ensure that the dataset is error-free, it is also critical to adopt measures to protect time series structures against data anomalies. This criteria stipulates that prior to the data being used in a model, concerns such as missing data and noise must be dealt with effectively. Some of the integrated approaches can be considered more robust, however, the final model has to take into account the rest, whose performance is affected by phenomenon such as a high noise level, outliers, gaps in the data (FS.01.02.02) and non-stationarity (FS.01.02.02)

	\item \textbf{Transformation of processed data} : Data transformations are commonly used to improve visual readability and rescale data instances into relative values in the event of normalization This transformation requirement dictates that the framework should handle these transformations.
	
\end{enumerate}

\paragraph*{Forecasting Framework (FR.02)} 
\begin{enumerate}
	\item \textbf{Integrate models}: Baseline methods can be used regardless of the problem context, but they only offer a bare minimum in terms of quality. On the other hand, more sophisticated approaches can outperform the simpler ones. This requirement implies combining those methods. The optimized models should support both Univariate Time Series (UTS) (FS.02.01.01) and Multivariate Time Series (MTS) (FS.02.01.02)
	
	\item \textbf{Implementation of Autonomous Approach Selection}: According to the features of the time series, the forecasting framework should be able to select the most appropriate method and determine the optimum parameters to match the time series' specifications (FS.02.02).
\end{enumerate}

When deciding on a model, one must take into consideration the likelihood of the model fitting. The performance of ARIMA and Long Short Term Memory (LSTM) techniques is projected to be lower when applied to small datasets. Requirements for how the autonomous approach selector manages the model's decision-making and parameterization are generated by the autonomous approach selector.

\begin{itemize}

	\item \textbf{Select the most fitting model}: This requirement (FR.02.01) states that approach selector should be able to determine the overall quality of the different approaches within a dataset, taking into consideration the characteristics of the dataset.
	
	\item \textbf{Select the most fitting parameter values}: Following the selection of a particular approach, it must be properly parametrized. This is important for all of the methods that are available. In case of k-NN, a few parameters should be anticipated (just the number of adjacent neighbours), whereas other approaches may warrant a greater number of parameters to minimize errors.

\end{itemize}

\paragraph*{Anomaly Detection framework (FS.03)} 
The requirements for the anomaly detection framework (FS.03) are identical to those for the framework for forecasting. The primary difference is the selection of the models the selector chooses from.

\paragraph*{Exposing framework operations through REST API}

A combination of the forecasting and anomaly detection frameworks constitute a time series analysis framework results. TSA framework also aggregate their data handling and model management potentials. Wrapping and exposing this framework through a REST API that offers all operations of the  forecasting and anomaly detection allows flexible integration into other services (FS.04).

\paragraph*{Performance Specifications}
Taking processing time into account is important, as some algorithms are more complex than others, which takes longer to complete. The Approach Selector should be time-bound in order to select the most appropriate approach.

\begin{enumerate}

	\item \textbf{Processing time for Approach Selection}:  It is evident that some Machine Learning (ML) techniques have drastically varying computational complexity since some techniques require much longer computation time. For the model to be usable, it is necessary that the model's implementation achieves a satisfactory performance threshold, according to this requirement. In this regard, a ratio that compares the quality of the algorithm with the complexity of the algorithm should be calculated and should be available for the user for adjustment.

	\item \textbf{Selection processing time}:  There are many different algorithms available to each of the selectors, which gives them the capabilities to deal with a variety of time series analysis challenges. The decision on the most appropriate method should have the least amount of impact on the overall execution time as possible.  In other words, this requirement should act as a ceiling to set the Autonomous Selector's execution time.
\end{enumerate}

\paragraph*{Quality Requirements}

A review of the literature reveals that different approaches, with different parametrization, show differing results for both prediction and anomaly detection practices. However, because some of the data sets collected already provide successful solutions,it is worthwhile to achieve a relative quality threshold, as explained in the below requirement.

\paragraph*{Optimum Threshold }

This requirement exploits datasets that are already in use and leverages already existing solutions to stipulate a quality threshold using metrics employed by the  solution. This is advantageous because the process can be adjusted to generate results of comparable quality to the M4-competitive processes and other solutions provided by other developers using other sources like the Irvine (UCI) University Repository.

\paragraph*{Technological Specifications}

The framework should be implemented using popular programming language like Java or Python and  Licenses for all other software used should enable its use without additional charge.

\section {General architecture of TSA framework}
\label{section_tsa_framework}

\begin{figure}
	
	\resizebox{0.9\linewidth}{!}{
		\centering
		
		\includegraphics[width=0.9\linewidth, height=0.8\linewidth]{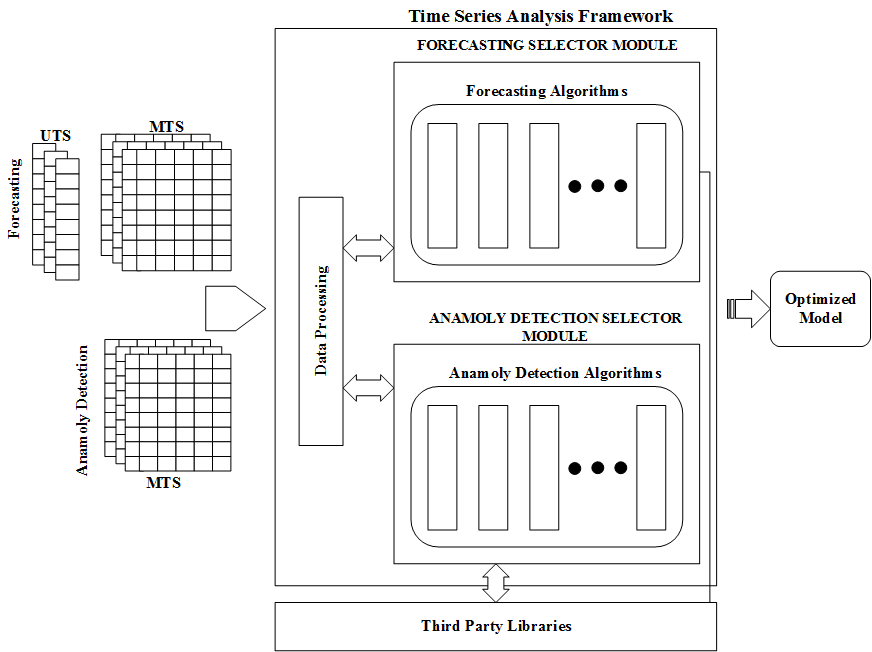}
		
	}
	\caption{\bf General architecture of Time Series Analysis framework.}
	\label{fig:tsa_genral_architecture}
\end{figure}

The primary modules of a typical TSA framework comprise of Forecasting Selector, Anomaly Detection Selector, Data Processing and Third Party Libraries.

This framework accepts time series as input, and then returns the best optimized and performing model, based on the type of analysis being performed (forecasting or anomaly detection). As shown in Figure \ref{fig:tsa_genral_architecture}, both UTS and MTS can be used in case of a forecasting and MTS, if anomaly detection is the preferred analysis.

The Forecasting Selector manages operations based on the data for which predictions are requested by the user. It maintains a number of predictive approaches which can be selected and correctly parameterised.

Like the Forecasting Selector, the Anomaly Detection Selector uses anomaly detection approaches, but applies them to Multivariate Time Series (MTS).

A variety of techniques are employed by the Data Processing module in order to ensure that data is correctly processed by the analysis selectors, while also improving the accuracy and performance of the approaches. This component also comes mainly from the third-party library supplied, which includes operations dealing with data characteristics. Scikit-learn \cite{pedregosa2011scikit}  and Pandas \cite{mckinney2011pandas} are such popular third-party libraries.  

In order to maximize performance while minimizing information loss, the Scikit-learn library is used for dataset feature reduction, which facilitates feature ranking, where the features with the most discriminating power are the most relevant.

Pandas \cite{mckinney2011pandas} is another powerful library that aids in data processing of Tabular data integrating other Python scientific computing libraries such as NumPy \cite{van2011numpy} and helps to eliminate inconsistencies that are incompatible with provided selection of approaches.

\section{Open source Tools for Time Series Analysis}
\label{section_tsa_tools}
A rising interest in machine learning and data mining has been seen in the past few years in many different fields.  In response to this surge in opportunities, there is a substantial increase in the number of machine learning packages, including both open source and proprietary. There are time series toolboxes that provide rich interfaces to specific models (ARIMA/filters \cite{seabold2010statsmodels}, neural networks \cite{alexandrov2019gluonts} ) or framework interfaces to individual time series models tasks (forecasting \cite{taylor2018forecasting} \cite{guecioueurpysf}, feature extraction \cite{kanter2015deep} \cite{tavenard2017tslearn} \cite{christ2018time}, Anomaly detection \cite{ayadi2017outlier} \cite{cook2019anomaly} \cite{wu2016survey}, annotation \cite{burns2018seglearn}, classification \cite{burns2018seglearn} \cite{tavenard2017tslearn}).

\subsection{Time Series Analysis Tools}

A number of effective toolboxes dedicated to time series analysis exist, which offer common interfaces to apply  modeling strategies of fitting, predicting, and hyper-parameter estimation and fine-tuning , as well as support for the building and tuning of composite models. More notably scikit-learn \cite{pedregosa2011scikit}, Orange \cite{demvsar2013orange} (Python), Weka \cite{holmes1994weka} \cite{hall2009weka} (Java), MLj \cite{blaom2020mlj} (Julia), and mlr \cite{bischl2016mlr}, mlr3 \cite{lang2019mlr3} or caret \cite{kuhn2015caret} (R).

Scikit-learn \cite{pedregosa2011scikit} is a Python module that incorporates a diverse set of cutting-edge machine learning algorithms for medium-scale supervised and unsupervised problems. This package is aimed at providing non-specialists with a high-level, general use of machine learning. Although most of the code is implemented in Python, there are also included C++ libraries with reference implementations of support vector machines (SVMs) \cite{cristianini2000introduction} and generalized linear models with compatible licenses. It incorporates compiled code for efficiency, with a lower reliance on NumPy  \cite{van2011numpy} and scipy \cite{jones2001scipy} only to facilitate easy distribution, and is primarily concerned with imperative programming.

Prophet \cite{taylor2018forecasting} is a Facebook-developed open-source forecasting library that is available in Python and R. It includes a single generic forecasting model based on the exponential-smoothing method family, but with a linear regression component as well.

Structural Time Series (STS) Tensorflow \cite{harvey1990forecasting} is a Google's open-source deep-learning framework recently announced. In Tensorflow Probability, STS is a part of Tensorflow's probabilistic modeling subpackage and is able to use its specialized machinery there.

Gluon Time Series (GluonTS) \cite{alexandrov2019gluonts} is an open-source modeling tool kit from Amazon Web Services.Based on MXNet's deep learning framework, GluonTS uses a diverse set of neural forecasting models. Due to its ease of use, GluonTS allows researchers to focus on developing models instead of dealing with tedious tasks like data input and model creation and testing.

Cesium \cite{naul2016cesium}, Seglearn \cite{burns2018seglearn} , tsfresh \cite{christ2018time}, and tslearn \cite{tavenard2017tslearn} are some notable Python libraries for processing time series and sequences with machine learning. These tools mainly assist time series data to feed models of other Python open source libraries such as Scikit-learn, pandasn numpy, scipy.

Cesium is a complete time series analysis platform composed of a Python library, along with a web-based front-end, and includes features such as the ability to examine data, apply machine learning algorithms, and provide simple, repeatable, and extendable results. The Seglearn package is a scikit-learn compatible open-source Python module that enables the efficient processing of  time series, sequence and contextual data for classification, regression, and forecasting issues. The 'tsfresh' (abbreviated for Time Series FeatuRe Extraction on basis of Scalable Hypothesis tests) is a python package that accelerates Time series feature engineering process by combining 63 time series characterization methods with the feature selection on the base of automatically defined testing hypotheses that by default calculate a total of 794 time series features. 'tslearn' is a Python package which includes preprocessor and functional extractor tools, as well as suitable models for the time series data (written as Scikit-learn compliant transformers).

Waikato Environment for Knowledge Analysis (Weka) (Version$\ge$3.7.3) \cite{hall2009weka} includes a dedicated time series analysis environment which enables the creation, evaluation, and visualization of forecasting models. The time series framework of Weka takes a machine learning/data mining approach to time series modeling by transforming data into a form that can be used by standard proposing learning algorithms. In Weka's graphical user interface, this environment takes the form of an online plugin tab and can be installed via the package manager. The information is encoded into additional input fields ( referred to as "lagged" variables), which removes the temporal ordering of individual input instances. The algorithms can model trends and seasonality by using a variety of other fields that are automatically computed. Any regression algorithms of Weka can be used to learn a model after data has been transformed. There are many methods that can be used to make predictions for a continuous target, such as  multiple linear regression, support vector machines for regression and model trees (decision trees with linear regression functions at the leaves).

The Orange \cite{demvsar2013orange} open-source machine learning and data mining framework is made up of scripting and visual programming elements. When it comes to visual interface development, Orange utilizes the concept of widgets; widgets, in this context, are essentially a graphical wrapper for complex data analysis algorithms that are developed in Orange and Python. Widgets communicate with one another through channels, and a collection of connected widgets is referred to as a schema. Either orange schemas can be set up in Python scripts or, preferably, designed in an application called Orange Canvas through visual programming. The 'As Time Series' widget in orange, reinterprets any data table as a time series data and allows to choose which data attribute represents time in the widget.

TSML \cite{palmes2020tsml} is an IBM package implemented in Julia  to process, classify, and forecast data for time series. TSML library provides common ML (Machine Learning) libraries from Python's Scikit-Learn, R's Caret and Julia Native MLs, to create complex assemblies that allow for robust time series predictions, clustering, and classification, in order to seamlessly integrate heterogeneous libraries.

The MLR3  \cite{lang2019mlr3} package for R language provides a framework for classification, regression, survival analysis, and clustering. It has a object-oriented uniform interface to more than 160 basic learners and contains feature selection, hyperparameter tuning, and ensemble creation meta-algorithms to improve and extend the capability of basic learners. By providing a single interface that allows for the extension and combination of current learners, as well as the intelligent selection and tuning of the most suited technique for a job, this approach effectively aids in the process of meta-learning. Natively enabled parallel high-performance computing through the 'parallelMap' package \cite{ParallelMap2021}  allows local multicore, socket, and MPI computation modes. Explicitly specifying which operations to parallelize is possible on compute clusters using 'BatchJobs' \cite{bischl2015batchjobs} package.  'mlr3temporal' package extends the mlr3 package framework by time-series forecasting and resampling methods.

The package named “Caret” \cite{kuhn2015caret} (abbreviated for classification and regression training) offers several tools for constructing predictive models in R, which have a vast variety of machine learning models available. The software includes features like as data splitting and pre-processing for projects in the early phases, as well as models for feature selection and resampling to help detect overfitting.

Some open source Feature-based time series foecasting tools in R language are tsfeatures \cite{hyndmanTsfeatures2021}, fforma \cite{montero2020fforma}, gratis \cite{kang2020gratis}, seer \cite{talagala2018meta}. 'tsfeatures' provides methods for extracting various features from time series data. The 'fforma' forecasting tools use a model combination approach to provide forecasting and can be used for model averaging or model selection. 'fforma' works by training a 'classifier,' or machine learning algorithm, how to select and combine forecast models. 
n feature-based time series forecasting, the 'gratis' package provides efficient algorithms for generating time series with a variety of controllable traits, which are helpful in creation of training data sets.. 
For forecast model selection, the "seer" package uses time series features and FFORMS (Feature-based FORecast Model Selection) to implement a novel framework.

\subsection{TSA Tools on Anomaly or Outlier Detection}

Environment for Developing KDD-Applications Supported by Index-Structures (ELKI for short) \cite{achtert2008elki} is a Java-based open source (AGPLv3) data mining software. ELKI's primary focus is algorithm research, with a particular emphasis on unsupervised methods in cluster analysis and outlier detection. ELKI can also integrate a geographic/geospatial information system (GIS) and a data mining system (DMS) into a single framework.

Anomaly Detection Extension in RapidMiner \cite{hofmann2016rapidminer} include popular unsupervised anomaly detection algorithms(implemented in Java). Using this tool, it is possible to find that data that is significantly different from normal, without having to label the data. RapidMiner's graphical user interface (GUI) can be used to create data mining processes that are made up of arbitrary nested operators that are described in XML files (GUI). Massive Online Analysis (MOA) \cite{bifet2010moa} is yet another Java-based software environment for developing algorithms and running experiments for online learning from changing data streams. MOA is intended to address the difficult problem of scaling up the implementation of cutting-edge algorithms to real-world dataset size and also supports bi-directional interaction with Weka \cite{hall2009weka},.

Machine learning toolboxes like the Shogun \cite{sonnenburg2010shogun} is built in C++ and have interfaces to Python and other languages like R, Octave and Java. This software includes support vector machines, dimensionality reduction and online learning as well as clustering and implemented kernels for numeric data analysis algorithms.

The Scikit-learn \cite{pedregosa2011scikit} library also has a set of machine learning tools that can be used for both novelty detection and outlier detection.  There are two methods for detecting anomalies: outliers and novelty detection. Outlier detection is also referred to as unsupervised anomaly detection, and novelty detection is referred to as semi-supervised anomaly detection in this context.

Python Outlier Detection (PyOD) \cite{zhao2019pyod} is a multivariate data outlier detection toolkit written in Python. Outlier ensembles and newly emerging deep learning models are included among the more than 20 detection algorithms that make up the toolkit. On top of PyOD, another acceleration framework called Scalable Unsupervised Outlier Detection (SUOD) \cite{zhao2020suod}  is built for large-scale unsupervised outlier detector training and prediction.

PyNomaly \cite{constantinou2018pynomaly} is a single algorithm implementation package in Python that implements a modified approach to Local Outlier Probabilities (LoOP) method. PyNomaly can be used for applications involving streaming data or when speedy calculations are required. Similary Jubatus \cite{hido2012jubatus} is a distributed processing framework and streaming machine learning library developed in C++ and support the Local Outlier Factor (LOF) method. The other two Python based packages implementing the Half Space Trees method are Creme \cite{halford2019creme} and scikit-multiflow \cite{montiel2018scikit}. The new library called River \cite{2020river} merges the features of Creme and scikit-multiflow.

The Anomaly Detection Toolkit (ADTK) \cite{arundo20202adtk} is a Python package that detects anomalies in time series using unsupervised / rule-based methods. This package provides a set of unified APIs for common detectors, transformers, and aggregators, as well as pipe classes that connect them into models. It also has some tools for processing and visualizing time series and anomaly events.

Python Streaming Anomaly Detection (PySAD) \cite{yilmaz2020pysad} is a Python framework for conducting anomaly detection experiments that provides a comprehensive set of tools. Over 15 online anomaly detection algorithms are available in the current (v0.1.1) version of PySAD as well as two alternative techniques for integrating PyOD detectors into the streaming environment.

TODS \cite{Lai_Zha_Wang_Xu_Zhao_Kumar_Chen_Zumkhawaka_Wan_Martinez_Hu_2021} is an automated Time Series Outlier Detection System designed for use in research and industry. Unlike other pipeline-building systems, TODS is extremely modular that enables the rapid construction of pipelines. TODS' \textit{primitive}  is an implementation of a function with hyper-parameters, and it serves as the foundation for all other components. Currently, TODS supports over 70 primitives, including functions for data manipulation and time series analysis, feature analysis, as well as algorithms for detection and a reinforcement module for training data. End-to-end outlier detection can be performed with the constructed pipeline by users who can freely build a pipeline using these primitives. Drag-and-drop functionality is provided by TODS with a  Graphical User Interface (GUI), that enables users to create flexible pipelines.

Skyline \cite{stanway2013etsy} is a real-time anomaly detection, time series analysis, and performance monitoring system designed to enable passive monitoring of metrics without the need to configure individual models/thresholds. It is intended for use in situations where a large number of high-resolution time series must be continuously monitored. Once a metrics stream is configured, Skyline automatically adds additional metrics for analysis.

Banpei \cite{tsurutaTsurubeeBanpei2021} is a Python package for anomaly detection that was created with the objective of establishing a real-time abnormality monitoring environment. Banpei has a function that corresponds to the streaming data in order to accomplish this goal. A simple monitoring system can also be built with the help of Bokeh \cite{BokehBokeh2021}, which is an excellent visualization library.

Telemanom \cite{hundman2018detecting} is a framework for detecting anomalies in multivariate time series data using Long Short-Term Memory (LSTMs). To find anomalies in multivariate sensor data, Telemanom uses vanilla LSTMs built with Keras/Tensorflow. These artificial neural networks (LSTMs) are taught to recognize common system behaviors by feeding them encoded command information and historical telemetry data. Every time step, predictions are generated, and predictions' errors represent deviations from expected behavior. Telemanom then employs a novel nonparametric, unsupervised approach to thresholding and identifying anomalous sequences of errors for Telemanom's algorithms.

For anomaly detection on time series data, DeepADoTS  \cite{fischer2019anomaly} is a bench-marking pipeline for many state-of-the-art deep learning approaches. On time series data, its main purpose is to create a benchmarking pipeline that can be used to compare the performance of various deep learning approaches.

In streaming, real-time applications, the Numenta Anomaly Benchmark (NAB) \cite{lavin2015evaluating} offers a fresh benchmark for testing algorithms for anomaly detection. It consists of more than 50 annotated real-world and artificial timeseries data files, as well as a revolutionary scoring method built specifically for real-time application development.

An access point to a list of 'R' packages that are mostly used for anomaly detection is provided by the 'CRAN Task View' \cite{hyndman2021cran}.  A collection of time series outlier detection methods is provided by the 'composits' package, which may be applied to compositional, multivariate, and univariate data. To find time series outliers, this package employs the R packages forecast, tsoutliers, otsad, and anomalize. The 'forecast' package provides some simple heuristic methods for identifying and correcting outliers. The 'tsoutliers' package is primarily intended for detecting outliers in time series. This package assist to discover patterns such as level shifts, temporary changes, seasonal level shifts, innovative outliers and additive outliers. The 'otsad' package contains a set of online defect (anomaly) detectors for time series that use prediction-based and window-based algorithms. It's capable of working in both stationary and non-stationary context.  'anomalize' is a  “tidy” workflow for detecting anomalies in data. An early detection package, referred to as a 'oddstream,' helps identify anomalous time series patterns in large collections of streaming time series data.

\section{Related Work}

\paragraph{Research topics and Application domains}
The field of time series analysis encompasses a wide range of research topics and applications domains. The literature contains numerous reviews on a variety of TSA topics, including forecasting ( \cite{hendikawati2020survey} \cite{mahalakshmi2016survey} \cite{panigrahi2020fuzzy} \cite{tealab2018time} ), anomaly detection(\cite{wu2016survey}, \cite{cook2019anomaly}, \cite{ayadi2017outlier} ). Some of the reviews focus on pattern recognition ( \cite{torkamani2017survey}, \cite{wang2019deep}) , dimensionality reduction(\cite{badhiye2018review})  and change point analysis ( \cite{sharma2016trend}, \cite{truong2020selective}, \cite{aminikhanghahi2017survey} ). Some reviews cover specific application domains such as IoT and Industry 4.0 ( \cite{lepenioti2020prescriptive}, \cite{mohammadi2018deep}, \cite{zhao2020review} ), health (\cite{zeger2006time}) and finance (\cite{SEZER2020106181}).

Few papers conduct systematic reviews of language-specific packages or libraries. There are several reviews of Python packages for various domains, such as data mining \cite{stanvcin2019overview}, topological data analysis \cite{ray2017survey}, or social media content scraping \cite{thivaharan2020survey}. With respect to R language, there are reviews related to packages for analyzing animal movement data in R \cite{joo2020navigating} , as well as hydrology-related R packages \cite{slater2019using}. The present article covers the open source Toolboxes for both Time Series Analysis and Anomaly Detection.

\section{Discussion}
\label{section_discussion}

\subsection{Tools for Time Series Analysis}

The tools for Time Series Analysis are largely divided into four groups based on the Graphical User Interface (GUI) functionality. First and foremost, tools that provide APIs but are primarily focused on the usage of a GUI that allows novices to quickly and simply apply machine learning algorithms (e.g Weka). Second, there are tools that focus exclusively on developing and presenting an useful and consistent API, implying that its intended users are capable of programming (e.g. scikit-learn and other compatible tools). The majority of tools explored in this article belong to this group. The third type of tools provide features both scripting and visual programming (e.g Orange). The fourth type of tools are intended to be used as command-line tools (and sometimes do not offer any type of API) (e.g sofia-ml).

With regard to time series learning capabilities, the tools shows different features. sktime \cite{loning2019sktime}  and tslearn \cite{tavenard2020tslearn} are generally devoted to the analysis of time series while tsfresh, cesium and seglearn concentrate on the extraction of time series statistical characteristics.Table \ref{tab:ToolsComparison1} shows the comparison of  time series learning capabilities of following tools:  cesium-ml, seglearn, sktime, tsfresh and tslearn. 

ceseium-ml (v0.9.6) and tsfresh (v0.11.1) provide multi-variable time series feature representation learning and are currently implementing more functionality than seglearn. Although the feature representation transformers are built as a pre-processing step that is distinct from the pipeline, the pipeline can nevertheless be used by other sklearn compatible tools. By design, this results in the inability to utilize end-to-end model selection. No obvious support exists for situations that affect a sequence or time series, or where deep learning models are integrated.

Time series specific classical methods for clustering, classification, and barycenter computation for time series with variable lengths have been implemented in the latest version of tslearn (v0.1.18.4). Feature representation learning, context learning or deep learning are not supported. Pyts \cite{faouzi2020pyts} is an entirely time series classification package for Python.


\begin{table}
\small
	\caption{
		 \bf 	Comparison of Time Series Learning features for Packages (versions): cesium (0.9.12), seglearn (1.2.2),  sktime (0.7.0), tsfresh (0.18.0), \& tslearn (0.5.2).
	 }
		\label{tab:ToolsComparison1}       
	\begin{adjustbox}{width=\linewidth}
		\begin{tabular}{ |  p{0.4\linewidth} |   p{0.12\linewidth}|   p{0.12\linewidth}  |   p{0.12\linewidth} |   p{0.12\linewidth} |  p{0.12\linewidth} |}
		\hline
			\bf  Feature &  \bf cesium-ml  & 	\bf seglearn & 	\bf sktime  &	\bf ts-fresh &    	\bf tslearn  \\  
		\hline

		Time series target &  \xmark & \cmark & \cmark & \xmark & \xmark \tabularnewline \cline{1-6}
		Sliding window segmentation & \xmark & \cmark & \cmark & \xmark  & \xmark   \tabularnewline \cline{1-6}
		Temporal folds   &  \xmark & \cmark & \cmark & \xmark  & \xmark \tabularnewline \cline{1-6}
		Context data  &  \cmark & \cmark & \cmark & \xmark  & \xmark \tabularnewline \cline{1-6}
		Multivariate time series  &  \cmark  & \cmark & \cmark & \cmark  & \cmark \tabularnewline \cline{1-6}
		Feature representation learning  &  \cmark  & \cmark & - & \xmark  & \xmark \tabularnewline \cline{1-6}
		Number of implemented features  & \centering 112 & \centering 27 & - & \centering 77  & \centering N/A  \tabularnewline \cline{1-6}
		Classification  &  \cmark & \cmark & \cmark & \cmark  & \cmark \tabularnewline \cline{1-6}\textbf{}
		Clustering  & \cmark & \cmark & \cmark & \cmark  & \cmark \tabularnewline \cline{1-6}
		Regression  &  \cmark & \cmark & \cmark & \cmark  & \cmark \tabularnewline \cline{1-6}
		Forecasting  & \cmark & \cmark & \cmark & \cmark & \xmark \tabularnewline \cline{1-6}
		Deep learning & \xmark & \cmark & \xmark & \xmark  & \xmark \tabularnewline \cline{1-6}
		sklearn compatible model selection  & \xmark &\cmark & \cmark & \xmark  & \xmark \tabularnewline \cline{1-6}

	\end{tabular}
	\end{adjustbox}
	

\end{table}


sktime and pyts stand out with their numbers of implementations of algorithms published in the literature. sktime provides more tree-based algorithms, while pyts is more focused on dictionary-based and image-based models.

Practitioners with a basic understanding of forecasting models but insufficient statistical knowledge should use Prophet if they have daily data \cite{januschowski2019open}. To handle data at a different granularity, you might need a different software program.In any circumstance, Prophet should be used for bench-marking. The Stan model code underlying Prophet provides greater programming flexibility for the more scientifically minded user.

The STS Tensorflow programming paradigm is beneficial for individuals who want to experiment with a programming paradigm that is connected with the TensorFlow environment \cite{januschowski2019open}. Tensorflow's momentum and speed of development mean that STS is bound to get better as the community around the tool grows and expands. At this point, it does not offer a complete forecasting tool package. The user still has to implement standard components like evaluation techniques and backtesting procedures, which is a task for any but the most skilled programmer.

For users undertaking research with deep learning models and developing scenarios for operational forecasting challenges, GluonTS can serve as the go-to tool set of choice \cite{januschowski2019open}. In terms of strategic and tactical forecasting, Prophet is better suited for users.

In many time series algorithms, the task of reducing a complex learning task to a simpler one is a common part of the process. Some of the frameworks allow to take advantage of reduction relationships that exist among various time series-related tasks to the fullest extent \cite{beygelzimer2008machine}. Many time series-related tasks, such as time series regression, multivariate (or panel) forecasting, but also forecasting and tabular (or cross-sectional) regression and time series annotation, anomaly detection have been shown to have reduction relationships \cite{bontempi2013machine}.  

Frameworks handle the time series dataset that can be either univariate, multivariate, or panel (sometimes referred to as longitudinal data), depending on the number and interrelation between time series variables and the number of instances for which each variable is observed. 

Frameworks implement standard interfaces for fitting, predicting and hyper-parameters, and support composite model development and tuning along with support for tabular (or cross-sectional) configuration. For fitting, forecasting, and hyper-parameters, almost all of the Toolboxes or libraries examined provide the same interface for composite model creation and adjustment. Toolbox capabilities, however, are still constrained when used outside of a cross-sectional context. Toolbox capabilities, however, are still constrained when used outside of a cross-sectional context. Toolbox capabilities, however, are still constrained when used outside of a cross-sectional context. The problem is that none of them have a dedicated forecasting API built in. Another class of toolboxes extends the capabilities of tabular toolboxes by giving functionality to handle certain parts of a time series modelling process, most notably feature extraction toolboxes such as Featuretools \cite{kanter2015deep}, tsfresh \cite{christ2018time} and hctsa \cite{fulcher2017hctsa}. Various smaller toolkits for specific reduction approaches, such as time series regression and forecasting, are also available, ranging from tabular toolboxes.

Additionally, there are several toolboxes devoted specifically to forecasting. However, the majority of them have significant drawbacks.
Forecast library \cite{hyndman2008automatic} in R is a widely used and comprehensive toolbox for forecasting. Forecast and its partner libraries provide significant capabilities for statistical and encapsulated machine learning algorithms, as well as for pre-processing, model selection, and assessment, when used in conjunction. Deep-learning models for probabilistic forecasting are also available in Python with GluonTS \cite{alexandrov2019gluonts} and it also interfaces with other programs such as forecast. The support for composite model generation, however, is restricted in both programs, and neither integrates with widely accessible machine learning libraries, such as scikit-Learn.

Certain model families can only be used with certain forecasting toolkits in Python. For example Time series analysis, including forecasting, can be done with statsmodels \cite{seabold2010statsmodels}, but it is limited to statistical models (e.g. ARIMA, exponential smoothing and state space models). However, pmdarima \cite{smith2017pmdarima} provides additional tools for seasonality testing, pre-processing and pipelining, but is limited to the ARIMA family of models. Likewise Prophet \cite{taylor2018forecasting} is restricted to generalised additive models whereas PyFlux \cite{taylor2016pyflux} only support generalised auto-regressive models(e.g. GAS, GARCH). 

In addition, a number of repositories, such as AtsPy \cite{snow2020atspy} and the Microsoft forecasting repository, gather and aggregate common forecasting models through interfaces to existing libraries and tools to simplify workflows, but none of them permit the construction of composite models.

\subsection{Discussion of TSA Tools on Anomaly detection}

In the current literature, existing frameworks are divided into two categories: (1) streaming frameworks and (2) anomaly detection frameworks. Unlike batch models, which may access all data with a finite number of instances during training, a streaming anomaly detection model takes data in the form of new instances and either retains the instance for a limited length and memory or trains itself immediately. Streaming frameworks are used for a variety of machine learning tasks using streaming data, including classification and regression, in addition to anomaly detection. Due to the requirement to maintain different methods for other activities at the same time, streaming frameworks have a limited number of specialized streaming anomaly detection methods. A pipeline made up of projectors, ensemblers, and probability calibrators can be used. In a streaming context, projectors project supplied information to a (potentially) lower dimensional space so that models may better distinguish abnormalities. Ensemblers create a score by combining scores from numerous models for a single instance. Using various techniques, probability calibrators transform target score into the probability of being abnormal.

Anomaly detection software is available in a variety of programming languages, including ELKI Data Mining, Massive Online Analysis (MOA) \cite{bifet2010moa} , and RapidMiner in Java. The 'CRAN Task View' \cite{hyndman2021cran} provides access to a list of 'R' packages dedicated to Anomaly detection like tsoutliers, anomalize and oddstream. There is also a large ecosystem of tools implemented in Python.

For anomaly detection, existing Toolboxes are mainly divided into three types: (1) standalone tools implementing single algorithm (like PyNomaly \cite{constantinou2018pynomaly} , Jubatus\cite{hido2012jubatus}), (2) part of a general larger framework that doesn't specifically cater to anomaly detection (Novelty and Outlier Detection in Scikit-learn \cite{pedregosa2011scikit}) and (3) toolbox dedicated to Anomaly detection (PyOD \cite{zhao2019pyod}, PySAD \cite{yilmaz2020pysad} etc). Also some frameworks focused on anomaly detection, such as PyOD and ADTK only target anomaly detection on batch data, whereas some framework ( like PySAD, Jubatus, MOA) targets anomaly detection on streaming data.

Table \ref{tab:ToolsComparison2} shows the comparison of  frameworks for streaming data and anomaly detection Packages of following tools:   PySAD (v0.1.1), PyOD (v0.9.5),  ADTK (v0.6.2), River (v0.8.0), scikit-multiflow (v0.5.3), Jubatus (v1.1.1) \& MOA (v2021.07).


\begin{table}
\small
	\caption{
		 \bf 	Comparison of some existing frameworks for streaming data and anomaly detection Packages (versions): PySAD (0.1.1), PyOD (0.9.5),  ADTK (0.6.2), River (0.8.0), scikit-multiflow (0.5.3), Jubatus (1.1.1) \& MOA (2021.07).
	 }
		\label{tab:ToolsComparison2}       
	\begin{adjustbox}{width=\linewidth}
		\begin{tabular}{ |  p{0.4\linewidth} |   p{0.12\linewidth}|   p{0.12\linewidth}  |   p{0.12\linewidth} |   p{0.12\linewidth} |  p{0.12\linewidth} |  p{0.12\linewidth} |  p{0.12\linewidth} | }
		\hline
			\bf  Feature &   \centering \bf PySAD  &  \centering	\bf PyOD &   \centering	\bf ADTK  &  \centering	\bf River &    \centering	\bf scikit-multiflow  &    \centering	\bf Jubatus  & \bf MOA \\  
		\hline

		Implementation Language & \centering Python & \centering Python & \centering Python  &  \centering Python & \centering Python & \centering C$++$ & \centering Java \tabularnewline \cline{1-8}
		\# of Models for Streaming Data & \centering 16 & \centering 0 & \centering 0 & \centering 2  & \centering 1 & \centering 1  & \centering 6 \tabularnewline \cline{1-8}
		Streaming  &   \cmark  & \xmark & \xmark & \cmark  & \cmark & \cmark & \cmark \tabularnewline \cline{1-8}
		Transformers  &  \cmark & \xmark & \cmark & \cmark  & \cmark & \cmark  & \cmark \tabularnewline \cline{1-8}
		Projectors  &  \cmark & \xmark & \xmark & \cmark  & \xmark & \xmark & \cmark \tabularnewline \cline{1-8}
		Ensemblers  &  \cmark  & \cmark & \cmark & \cmark  & \cmark & \xmark & \cmark \tabularnewline \cline{1-8}
		Calibrators  &  \cmark  & \xmark & \xmark & \xmark  & \xmark & \xmark & \xmark \tabularnewline \cline{1-8}

	\end{tabular}
	\end{adjustbox}
	

\end{table}


Out of the compared frameworks some support  streaming data (PySAD, River, scikit-multiflow, Jubatus and MOA), while others support batch data only (ADTK, PyOD). Except Jubatus, all packages support Ensemble. Projector component is supported in only River, MOA and PySAD. The PySAD has unsupervised probability calibrators that convert anomaly scores to probabilities. Anomaly scores are rarely interpretable in a probabilistic manner, and they cannot be easily transformed into anomalousness decisions.

\subsection{Overall analysis of tools}

This research work carried out the analysis of 60 tools which provide different time analysis tasks (See Table \ref{tab:all_tools_summary_tsa_v1_1}). The analysis found that 32 tools explicitly providing forecasting functinality (TSK1). Forecasting is by far the most frequently implemented task by most of the tools. Classification tasks (TSK2) are implemented by 14 Tools. 13 Tools are classsfied to provide clustering methods (TSK3).  Anomaly Detection (TSK4) is the second most frequently implemented task by 21 tools. Five Tools also provided segmentation methods (TSK5). Eight Tools are classified under the category pattern recognition (TSK6), consisting of both indexing and motif discovery tasks. Eight Tools are identified to provide change point detection (TSK7) functionality.

In terms of providing the data preparation capabilities, only seven tools are identified to explicitly provide functionality of dimensionality reduction methods (DP1). The facility of imputing missing values (DP2) exist in 23 Tools. 24 Tools support the Decomposition methods (DP3) such as decomposing time series into trends, seasonal components or frequency components.  Explicit generic transformation and features generation methods (DP4) are implemented in 24 Tools. The are limited number of tools (only nine) that support the functionality of providing methods for computing similarity measures (DP5). It has to be noted that there are less tools providing the data preparation capabilities of DP1 and DP5. There is no significant difference, in terms of number of tools providing other data preparation capabilities DP2, DP3 and DP4.

During the analysis on tools that provide generation of synthetic time series data (DS1), it is found that 22 tools provide this feature, while 31 tools contain specific data-sets for time series. The vast majority of these tools also include metric to evaluate the output generated. After the analysis it is found that there are 27 tools that explicitly provide methods for model selection (EVL1). Evaluation metric and statistical tests (EVL2) are included in 41 tools. Standalone visualization methods (EVL3) are provided in 39 tools.

\section{Conclusion }
\label{section_conclusion_rec}

This paper describes a typical Time Series Analysis (TSA) framework with architecture and listed out the main features of TSA framework. Further this article conducted a comprehensive review of open source tools for time series analysis. The tools are categorized based on the criteria of analysis tasks completed, data preparation methods employed, and evaluation methods for results generated. However this review work does not carry out evaluation of implementations or results for these tools, such as those on benchmark dataset. Overall, this article considered 60 time series analysis tools and 32 tools provided forecasting module and 21 packages include anomaly detection. A list and comparison of Tools for TSA which provide automation in time series analysis, for forecasting and anomaly detection are provided. Majority of the tools in ecosystem exists in Python, since it is the programming languages of choice for data scientists. The overall review activity showed that the existing Tool ecosystem is fragmented, with several specialized tool kits for Time series analysis  but no overall framework, making it difficult to understand, utilize, and inter operate with others tools.


\bibliographystyle{unsrtnat}
\bibliography{timeseries_analysis_frameworks}

\appendix

\begin{landscape}
	\label{Appendix_requirments}
	
 \begin{longtable}{| p{0.1\linewidth} |  p{0.2\linewidth} |  p{0.5\linewidth} |   p{0.1\linewidth} |}
 	\caption{Technical Specifications of a Time Series Analysis Framework.}
 	\label{tab:requirments} 
 	\\
	\bf  ID & \bf Requirement Title & 	\bf Description & \bf Dependencies \\  
	\hline

	\multicolumn{4}{|c|}{ \bf Functional Specifications} \\
	\hline
	FS.01	& Build capabilities for data processing 	& Data preparation and processing should be handled within the framework.	& -
\\
	\hline
	FS.02	& Build a framework for Forecasting 	& Provide an interface for forecast operations that manages different forecast approaches for different time series and selects the most suited, properly parametersed approach.	& FS.01 \\
	\hline
	FS.03	& Build a Anomaly Detection framework	& IIncorprate an anomaly detection framework that manages different anomaly detection approaches and performances,\& outputs the most appropriate, based on the features of the input dataset.	& FS.01 \\

	\hline
	FS.04	& Exposing framework operations through REST API	&  The framework should be available as REST APIs wrapped in an API interface. & - \\
	\hline
	
	\multicolumn{4}{|c|}{ \bf Specifications for Implementation of data processing techniques} \\
	\hline
	
	FS.01.01	& Physical data processing characteristics	& Raw data should be organized into a tabular format,by the framework, so that the approaches can understand. & -
\\
	\hline
	
	FS.01.02	& Processing time series data characteristics & The implementation of features that are specific to time series data should be provided. & FS.01.01
\\
	\hline
	
	FS.01.03	& Transformation of processed data & Allow techniques for dealing with both conventional statistical data transformations and machine learning transformations. & FS.01.01, FS.01.02 \\
	\hline
	
	\multicolumn{4}{|c|}{ \bf Specifications for Processing time series data characteristics} \\
	\hline
	
	FS.01.02.01	& Non-stationarity is addressed &  Non-stationarity characerstics like Seasonality and trend should be detected and removed using various data processing techniques.	& -
\\
	\hline
	FS.01.02.02	 & Detect and manage missing data & When there are missing data occurances, interpolation should be used to mask these instances.	& - \\
	\hline
	
	\multicolumn{4}{|c|}{ \bf Specifications to Build a Forecasting framework (FS.02)} \\
	\hline
	
	FS.02.01	& Integrate models	& The framework should incorporate a variety of specialized approaches, the selection of which should be determined by the data.	& -
\\
	\hline
	
	FS.02.02	& Implementation of Autonomous Approach Selection & The implementation should include support for automating the decision process of optimal forecasting approach which includes the step of model parameterization.	& FS.02.01\\
	\hline

	\multicolumn{4}{|c|}{ \bf Specifications for models’ integration (FS.02.01) )} \\
	\hline
	FS.02.01.01	& Integrate models for UTS	& Optimized models that have been implemented should be suitable for Univariate Time Series (UTS).	& -
\\
	\hline
	FS.02.01.02	& Integrate models for MTS	& Optimized models that have been implemented should be suitable for Multivariate Time Series (MTS)	& -\\
	\hline
	
	\multicolumn{4}{|c|}{ \bf Specifications for implementation of Autonomous Approach Selection (FS.02.02) } \\
	\hline
	FS.02.02.01	 & Select   the   most   fitting model	& Given the dataset and its characteristics, the approach selector can evaluate approaches available in terms of their quality.	& -
 \\
	\hline
	FS.02.02.02	& Select the most fitting parameter values	& Once an approach is adopted, the selector should choose an approach to parametrize the results so as to maximize the quality of the results. 	& FS.02.03.01  \\
	\hline

	\multicolumn{4}{|c|}{ \bf Performance Specifications } \\
	\hline
	PS.01	& Processing time for Approach Selection	& The Approach Selector should take into consideration the processing time for each approach.	& -
 \\
	\hline
	PS.02	& Selection processing time	 & The Approach Selector should be implemented in such a way that it minimizes the time required for selection.	& -  \\
	\hline

	\multicolumn{4}{|c|}{ \bf Quality Specifications } \\
	\hline
	QS.01	& Optimum Threshold	& According to a given benchmark, the framework should define a minimum accuracy. & -  \\
	\hline

	\hline

\end{longtable}

\end{landscape}

\begin{landscape}
\label{Appendix_table_all_tools_summary_tsa_v1_1}
\small
 \begin{longtable}{| m{0.17\linewidth} 
							  |  m{0.03\linewidth} |  m{0.03\linewidth} |   m{0.03\linewidth}  |   m{0.03\linewidth} |   m{0.03\linewidth} |   m{0.03\linewidth}  |   m{0.03\linewidth} 
 							  |   m{0.03\linewidth} |    m{0.03\linewidth} |   m{0.03\linewidth} |   m{0.03\linewidth} | m{0.03\linewidth} 
 							  |   m{0.03\linewidth} |  m{0.03\linewidth} |  m{0.03\linewidth} 
 							  |   m{0.03\linewidth} |  m{0.03\linewidth} | 
 						  }
 	\caption{Summary of Tools for TSA.}
 	\label{tab:all_tools_summary_tsa_v1_1} 
 	
 	\\ \hline
 	

    \bf  &  \multicolumn{7}{|c|}{\textbf{Tasks}}
	& \multicolumn{5}{|c|}{\textbf{Data Preparation}}
	& \multicolumn{2}{|c|}{\textbf{Dataset}}
	&  \multicolumn{3}{|c|}{\textbf{Evaluation}} \\  
	\hline		 
		 	
	\bf Tool Name   & \bf TSK1 & 	\bf TSK2 & \bf TSK3 & \bf TSK4 & \bf TSK5 & \bf TSK6 & \bf TSK7
	& \bf DP1 & \bf DP2 & \bf DP3 & \bf DP4 & \bf DP5
	& \bf DS1  & \bf DS2
	& \bf EVL1  & \bf EVL2 & \bf EVL3 \\  
	\hline
	\endfirsthead
   \multicolumn{18}{@{}l}{\ldots Continued from Previous Page}\\\hline
   
   \bf Tool Name   & \bf TSK1 & 	\bf TSK2 & \bf TSK3 & \bf TSK4 & \bf TSK5 & \bf TSK6 & \bf TSK7
   & \bf DP1 & \bf DP2 & \bf DP3 & \bf DP4 & \bf DP5
   & \bf DS1  & \bf DS2
   & \bf EVL1  & \bf EVL2 & \bf EVL3 \\  
   \hline
   \endhead 
   \hline
   \multicolumn{18}{r@{}}{Continued on Next Page \ldots}\\
   \endfoot
   
   \multicolumn{18}{|D{1\linewidth}|}{{
   		\begin{flushleft}
   		\bf{Note: TSA Tasks: TSK1 - Forecasting methods (FS.02), TSK2 - Classification methods, TSK3 - Clustering methods, TSK4 - Anomaly Detection methods (FS.03), TSK5 - Segmentation methods, TSK6 - Pattern Recognition, TSK7 - Change Point Detection. Data Preparation Modules: DP1 - Dimensionality reduction methods, DP2 - Missing values imputation methods (FS.01.02.02), DP3 - Decomposition methods (e.g., decomposing time series into trends, seasonal components, or frequency components) (FS.01.02.01), DP4 - preprocessing, DP5 - Similarity measures. Datasets: D1 - Generating synthetic time series data, D2- Providing access to time series datasets. Evaluation Components: EVL1 - Methods for model selection, hyperparameter search, or feature selection, EVL2 - Providing evaluation metrics and statistical tests, EVL3 - Providing visualization. \moduleexist \space - Module Exists. Primary Implementation language: \pythonprimary - Python, \rprimary - R, \javaprimary - Java, \juliaprimary - Julia, \cppprimary - C++. } 
   		\end{flushleft}
    }} \\ \hline
   \endlastfoot


	
	
    ADTK \pythonprimary &  \text{}  &  \text{}   &   \text{}   &   \text{\moduleexist}   &   \text{}  &    \text{\moduleexist}  &  \text{}   
	&  \text{}   &  \text{}   &  \text{\moduleexist} &  \text{}  &  \text{} 
	&  \text{}  &  \text{} 
	&  \text{}   &  \text{}  & \text{\moduleexist} \\
	\hline

    arch \pythonprimary &  \text{\moduleexist}  &  \text{}   &   \text{}   &   \text{}   &   \text{}  &    \text{}  &  \text{}   
	&  \text{}   &  \text{}   &  \text{} &  \text{}  &  \text{} 
	&  \text{\moduleexist}  &  \text{\moduleexist} 
	&  \text{}   &  \text{\moduleexist}  & \text{} \\
	\hline
	
	atspy \pythonprimary  &  \text{\moduleexist}  &  \text{}   &   \text{}   &   \text{}   &   \text{}  &    \text{}  &  \text{}   
	&  \text{}   &  \text{\moduleexist}   &  \text{\moduleexist} &  \text{\moduleexist}  &  \text{} 
	&  \text{}  &  \text{} 
	&  \text{\textbf{\moduleexist}}   &  \text{\moduleexist}  & \text{\moduleexist} \\
	\hline
	
	banpei \pythonprimary &  \text{}  &  \text{}   &   \text{}   &   \text{\moduleexist}   &   \text{}  &    \text{}  &  \text{\moduleexist}   
	&  \text{}   &  \text{}   &  \text{} &  \text{}  &  \text{} 
	&  \text{}  &  \text{} 
	&  \text{}   &  \text{}  & \text{} \\
	\hline
	
	cesium \pythonprimary &  \text{}  &  \text{}   &   \text{}   &   \text{}   &   \text{}  &    \text{}  &  \text{}   
	&  \text{}   &  \text{}   &  \text{} &  \text{\moduleexist}  &  \text{} 
	&  \text{}  &  \text{} 
	&  \text{}   &  \text{}  & \text{} \\
	\hline
	
	Caret in R \rprimary &  \text{\moduleexist}  &  \text{}   &   \text{}   &   \text{}   &   \text{}  &    \text{}  &  \text{}   
	&  \text{}   &  \text{\moduleexist}   &  \text{} &  \text{}  &  \text{} 
	&  \text{}  &  \text{\moduleexist} 
	&  \text{\moduleexist}   &  \text{\moduleexist}  & \text{\moduleexist} \\
	\hline
	
	CRAN Task View  \rprimary &  \text{\moduleexist}  &  \text{}   &   \text{\moduleexist}   &   \text{\moduleexist}   &   \text{}  &    \text{\moduleexist}  &  \text{\moduleexist}   
	&  \text{\moduleexist}   &  \text{\moduleexist}   &  \text{\moduleexist} &  \text{}  &  \text{\moduleexist} 
	&  \text{\moduleexist}  &  \text{\moduleexist} 
	&  \text{\moduleexist}   &  \text{\moduleexist}  & \text{\moduleexist} \\
	\hline
	
	darts \pythonprimary &  \text{\moduleexist}  &  \text{}   &   \text{}   &   \text{}   &   \text{}  &    \text{}  &  \text{}   
	&  \text{}   &  \text{\moduleexist}   &  \text{\moduleexist} &  \text{\moduleexist}  &  \text{} 
	&  \text{\moduleexist}  &  \text{\moduleexist} 
	&  \text{}   &  \text{\moduleexist}  & \text{\moduleexist} \\
	\hline
	
	DeepADoTS \pythonprimary &  \text{}  &  \text{}   &   \text{}   &   \text{\moduleexist}   &   \text{}  &    \text{}  &  \text{}   
	&  \text{}   &  \text{}   &  \text{} &  \text{}  &  \text{} 
	&  \text{}  &  \text{} 
	&  \text{}   &  \text{\moduleexist}  & \text{\moduleexist} \\
	\hline
	
	deeptime \pythonprimary &  \text{\moduleexist}  &  \text{}   &   \text{\moduleexist}   &   \text{}   &   \text{}  &    \text{}  &  \text{}   
	&  \text{\moduleexist}   &  \text{}   &  \text{\moduleexist} &  \text{\moduleexist}  &  \text{} 
	&  \text{\moduleexist}  &  \text{} 
	&  \text{}   &  \text{\moduleexist}  & \text{\moduleexist} \\
	\hline
	
	deltapy \pythonprimary  &  \text{\moduleexist}  &  \text{\moduleexist}   &   \text{}   &   \text{}   &   \text{}  &    \text{}  &  \text{}   
	&  \text{\moduleexist}   &  \text{}   &  \text{\moduleexist} &  \text{\moduleexist}  &  \text{\moduleexist} 
	&  \text{}  &  \text{} 
	&  \text{}   &  \text{}  & \text{} \\
	\hline
	
	dtaidistance \pythonprimary &  \text{}  &  \text{}   &   \text{\moduleexist}   &   \text{}   &   \text{}  &    \text{}  &  \text{}   
	&  \text{}   &  \text{}   &  \text{} &  \text{}  &  \text{\moduleexist} 
	&  \text{}  &  \text{} 
	&  \text{}   &  \text{}  & \text{\moduleexist} \\
	\hline
	
	EMD-signal \pythonprimary &  \text{}  &  \text{}   &   \text{}   &   \text{}   &   \text{}  &    \text{}  &  \text{}   
	&  \text{}   &  \text{}   &  \text{\moduleexist} &  \text{}  &  \text{} 
	&  \text{}  &  \text{} 
	&  \text{}   &  \text{}  & \text{\moduleexist} \\
	\hline
	
	ELKI Data Mining \javaprimary &  \text{}  &  \text{}   &   \text{\moduleexist}   &   \text{\moduleexist}   &   \text{}  &    \text{}  &  \text{}   
	&  \text{}   &  \text{}   &  \text{} &  \text{}  &  \text{} 
	&  \text{\moduleexist}  &  \text{\moduleexist} 
	&  \text{\moduleexist}   &  \text{}  & \text{} \\
	\hline
	
	flood-forecast \pythonprimary &  \text{\moduleexist}  &  \text{}   &   \text{}   &   \text{}   &   \text{}  &    \text{}  &  \text{}   
	&  \text{}   &  \text{\moduleexist}   &  \text{} &  \text{\moduleexist}  &  \text{} 
	&  \text{}  &  \text{} 
	&  \text{}   &  \text{\moduleexist}  & \text{\moduleexist} \\
	\hline
	
	gluonts \pythonprimary &  \text{\moduleexist}  &  \text{}   &   \text{}   &   \text{}   &   \text{}  &    \text{}  &  \text{}   
	&  \text{}   &  \text{\moduleexist}   &  \text{} &  \text{\moduleexist}  &  \text{} 
	&  \text{\moduleexist}  &  \text{\moduleexist} 
	&  \text{}   &  \text{\moduleexist}  & \text{\moduleexist} \\
	\hline
	
	hcrystalball \pythonprimary &  \text{\moduleexist}  &  \text{}   &   \text{}   &   \text{}   &   \text{}  &    \text{}  &  \text{}   
	&  \text{}   &  \text{}   &  \text{} &  \text{}  &  \text{} 
	&  \text{\moduleexist}  &  \text{\moduleexist} 
	&  \text{\moduleexist}   &  \text{\moduleexist}  & \text{\moduleexist} \\
	\hline
	
	hmmlearn \pythonprimary  &  \text{\moduleexist}  &  \text{}   &   \text{}   &   \text{}   &   \text{}  &    \text{}  &  \text{}   
	&  \text{}   &  \text{\moduleexist}   &  \text{} &  \text{\moduleexist}  &  \text{} 
	&  \text{\moduleexist}  &  \text{} 
	&  \text{}   &  \text{}  & \text{} \\
	\hline
	
	hypertools \pythonprimary &  \text{}  &  \text{}   &   \text{\moduleexist}   &   \text{}   &   \text{}  &    \text{}  &  \text{}   
	&  \text{\moduleexist}   &  \text{\moduleexist}   &  \text{} &  \text{\moduleexist}  &  \text{} 
	&  \text{}  &  \text{} 
	&  \text{}   &  \text{}  & \text{\moduleexist} \\
	\hline
	
	Jubatus \cppprimary &  \text{\moduleexist}  &  \text{\moduleexist}   &   \text{\moduleexist}   &   \text{\moduleexist}   &   \text{}  &    \text{}  &  \text{}   
	&  \text{}   &  \text{}   &  \text{} &  \text{}  &  \text{} 
	&  \text{}  &  \text{} 
	&  \text{}   &  \text{\moduleexist}  & \text{\moduleexist} \\
	\hline
	
	linearmodels \pythonprimary &  \text{}  &  \text{}   &   \text{}   &   \text{}   &   \text{}  &    \text{}  &  \text{}   
	&  \text{}   &  \text{}   &  \text{} &  \text{}  &  \text{} 
	&  \text{}  &  \text{\moduleexist} 
	&  \text{}   &  \text{\moduleexist}  & \text{} \\
	\hline
	
	luminaire \pythonprimary &  \text{\moduleexist}  &  \text{}   &   \text{}   &   \text{\moduleexist}   &   \text{}  &    \text{}  &  \text{\moduleexist}   
	&  \text{}   &  \text{\moduleexist}   &  \text{\moduleexist} &  \text{\moduleexist}  &  \text{} 
	&  \text{}  &  \text{} 
	&  \text{\moduleexist}   &  \text{}  & \text{} \\
	\hline
	
	Massive Online Analysis \javaprimary &  \text{\moduleexist}  &  \text{\moduleexist}   &   \text{\moduleexist}   &   \text{\moduleexist}   &   \text{}  &    \text{}  &  \text{}   
	&  \text{}   &  \text{\moduleexist}   &  \text{\moduleexist} &  \text{\moduleexist}  &  \text{} 
	&  \text{}  &  \text{\moduleexist} 
	&  \text{\moduleexist}   &  \text{\moduleexist}  & \text{\moduleexist} \\
	\hline
	
	matrixprofile \pythonprimary &  \text{}  &  \text{}   &   \text{\moduleexist}   &   \text{\moduleexist}   &   \text{\moduleexist}  &    \text{\moduleexist}  &  \text{}   
	&  \text{}   &  \text{\moduleexist}   &  \text{} &  \text{}  &  \text{\moduleexist} 
	&  \text{}  &  \text{\moduleexist} 
	&  \text{}   &  \text{}  & \text{\moduleexist} \\
	\hline
	
	mcfly \pythonprimary &  \text{}  &  \text{\moduleexist}   &   \text{}   &   \text{}   &   \text{}  &    \text{}  &  \text{}   
	&  \text{}   &  \text{}   &  \text{} &  \text{}  &  \text{} 
	&  \text{}  &  \text{} 
	&  \text{\moduleexist}   &  \text{}  & \text{\moduleexist} \\
	\hline
	
	
	
	MLR3 \rprimary  &  \text{\moduleexist}  &  \text{\moduleexist}   &   \text{\moduleexist}   &   \text{}   &   \text{}  &    \text{}  &  \text{}   
	&  \text{}   &  \text{\moduleexist}   &  \text{\moduleexist} &  \text{}  &  \text{} 
	&  \text{\moduleexist}  &  \text{\moduleexist} 
	&  \text{\moduleexist}   &  \text{\moduleexist}  & \text{\moduleexist} \\
	\hline
	
	neuralprophet \pythonprimary &  \text{\moduleexist}  &  \text{}   &   \text{}   &   \text{}   &   \text{}  &    \text{}  &  \text{}   
	&  \text{}   &  \text{\moduleexist}   &  \text{\moduleexist} &  \text{\moduleexist}  &  \text{} 
	&  \text{}  &  \text{} 
	&  \text{}   &  \text{}  & \text{\moduleexist} \\
	\hline
	
	nolds \pythonprimary &  \text{}  &  \text{}   &   \text{}   &   \text{}   &   \text{}  &    \text{}  &  \text{}   
	&  \text{}   &  \text{\moduleexist}   &  \text{} &  \text{}  &  \text{} 
	&  \text{\moduleexist}  &  \text{\moduleexist} 
	&  \text{}   &  \text{\moduleexist}  & \text{} \\
	\hline
	
	Numenta AB \pythonprimary  &  \text{}  &  \text{}   &   \text{}   &   \text{\moduleexist}   &   \text{}  &    \text{}  &  \text{\moduleexist}   
	&  \text{}   &  \text{}   &  \text{} &  \text{}  &  \text{} 
	&  \text{}  &  \text{\moduleexist} 
	&  \text{\moduleexist}   &  \text{\moduleexist}  & \text{} \\
	\hline
	
	Orange \pythonprimary &  \text{\moduleexist}  &  \text{}   &   \text{}   &   \text{}   &   \text{}  &    \text{}  &  \text{}   
	&  \text{}   &  \text{}   &  \text{\moduleexist} &  \text{}  &  \text{} 
	&  \text{}  &  \text{} 
	&  \text{}   &  \text{\moduleexist}  & \text{\moduleexist} \\
	\hline
	
	pmdarima \pythonprimary &  \text{\moduleexist}  &  \text{}   &   \text{}   &   \text{}   &   \text{}  &    \text{}  &  \text{}   
	&  \text{}   &  \text{}   &  \text{\moduleexist} &  \text{\moduleexist}  &  \text{} 
	&  \text{}  &  \text{\moduleexist} 
	&  \text{\moduleexist}   &  \text{\moduleexist}  & \text{\moduleexist} \\
	\hline
	
	prophet \pythonprimary &  \text{\moduleexist}  &  \text{}   &   \text{}   &   \text{}   &   \text{}  &    \text{}  &  \text{\moduleexist}   
	&  \text{}   &  \text{}   &  \text{\moduleexist} &  \text{}  &  \text{} 
	&  \text{}  &  \text{} 
	&  \text{}   &  \text{\moduleexist}  & \text{\moduleexist} \\
	\hline
	
	pyaf \pythonprimary &  \text{\moduleexist}  &  \text{}   &   \text{}   &   \text{}   &   \text{}  &    \text{}  &  \text{}   
	&  \text{}   &  \text{\moduleexist}   &  \text{\moduleexist} &  \text{\moduleexist}  &  \text{} 
	&  \text{}  &  \text{\moduleexist} 
	&  \text{\moduleexist}   &  \text{\moduleexist}  & \text{\moduleexist} \\
	\hline
	
	pycwt \pythonprimary &  \text{}  &  \text{}   &   \text{}   &   \text{}   &   \text{}  &    \text{}  &  \text{}   
	&  \text{}   &  \text{}   &  \text{\moduleexist} &  \text{}  &  \text{} 
	&  \text{}  &  \text{\moduleexist} 
	&  \text{}   &  \text{}  & \text{} \\
	\hline
	
	pydlm \pythonprimary &  \text{\moduleexist}  &  \text{}   &   \text{}   &   \text{}   &   \text{}  &    \text{}  &  \text{}   
	&  \text{}   &  \text{}   &  \text{\moduleexist} &  \text{}  &  \text{} 
	&  \text{}  &  \text{} 
	&  \text{\moduleexist}   &  \text{\moduleexist}  & \text{\moduleexist} \\
	\hline
	
	pyFTS \pythonprimary &  \text{\moduleexist}  &  \text{}   &   \text{}   &   \text{}   &   \text{}  &    \text{}  &  \text{}   
	&  \text{}   &  \text{}   &  \text{} &  \text{\moduleexist}  &  \text{} 
	&  \text{\moduleexist}  &  \text{\moduleexist} 
	&  \text{\moduleexist}   &  \text{\moduleexist}  & \text{\moduleexist} \\
	\hline
	
	PyNomaly \pythonprimary  &  \text{}  &  \text{}   &   \text{}   &   \text{\moduleexist}   &   \text{}  &    \text{}  &  \text{}   
	&  \text{}   &  \text{}   &  \text{} &  \text{}  &  \text{} 
	&  \text{}  &  \text{} 
	&  \text{}   &  \text{\moduleexist}  & \text{} \\
	\hline
	
	PySAD \pythonprimary  &  \text{}  &  \text{}   &   \text{}   &   \text{\moduleexist}   &   \text{}  &    \text{}  &  \text{}   
	&  \text{}   &  \text{}   &  \text{} &  \text{}  &  \text{} 
	&  \text{}  &  \text{} 
	&  \text{\moduleexist}   &  \text{\moduleexist}  & \text{} \\
	\hline
	
	PyOD \pythonprimary  &  \text{}  &  \text{}   &   \text{}   &   \text{\moduleexist}   &   \text{}  &    \text{}  &  \text{}   
	&  \text{}   &  \text{}   &  \text{} &  \text{}  &  \text{} 
	&  \text{\moduleexist}  &  \text{} 
	&  \text{\moduleexist}   &  \text{\moduleexist}  & \text{\moduleexist} \\
	\hline
	
	pyodds \pythonprimary  &  \text{}  &  \text{}   &   \text{}   &   \text{\moduleexist}   &   \text{}  &    \text{}  &  \text{}   
	&  \text{}   &  \text{}   &  \text{} &  \text{}  &  \text{} 
	&  \text{}  &  \text{} 
	&  \text{\moduleexist}   &  \text{\moduleexist}  & \text{\moduleexist} \\
	\hline
	
	
	pytorchts \pythonprimary  &  \text{\moduleexist}  &  \text{}   &   \text{}   &   \text{}   &   \text{}  &    \text{}  &  \text{}   
	&  \text{}   &  \text{\moduleexist}   &  \text{} &  \text{\moduleexist}  &  \text{} 
	&  \text{\moduleexist}  &  \text{\moduleexist} 
	&  \text{}   &  \text{}  & \text{\moduleexist} \\
	\hline
	
	pyts \pythonprimary &  \text{}  &  \text{\moduleexist}   &   \text{}   &   \text{}   &   \text{\moduleexist}  &    \text{\moduleexist}  &  \text{}   
	&  \text{}   &  \text{\moduleexist}   &  \text{\moduleexist} &  \text{\moduleexist}  &  \text{\moduleexist} 
	&  \text{\moduleexist}  &  \text{\moduleexist} 
	&  \text{}   &  \text{\moduleexist}  & \text{\moduleexist} \\
	\hline
	
	PyWavelets \pythonprimary &  \text{}  &  \text{}   &   \text{}   &   \text{}   &   \text{}  &    \text{}  &  \text{}   
	&  \text{}   &  \text{}   &  \text{\moduleexist} &  \text{\moduleexist}  &  \text{} 
	&  \text{\moduleexist}  &  \text{\moduleexist} 
	&  \text{}   &  \text{}  & \text{} \\
	\hline
	
	ruptures \pythonprimary &  \text{}  &  \text{}   &   \text{}   &   \text{}   &   \text{}  &    \text{}  &  \text{\moduleexist}   
	&  \text{}   &  \text{}   &  \text{} &  \text{}  &  \text{} 
	&  \text{}  &  \text{} 
	&  \text{\moduleexist}   &  \text{\moduleexist}  & \text{\moduleexist} \\
	\hline
	
	
	
	scikit-multiflow \pythonprimary &  \text{}  &  \text{\moduleexist}   &   \text{}   &   \text{\moduleexist}   &   \text{}  &    \text{}  &  \text{\moduleexist}   
	&  \text{}   &  \text{\moduleexist}   &  \text{} &  \text{}  &  \text{\moduleexist} 
	&  \text{\moduleexist}  &  \text{} 
	&  \text{}   &  \text{\moduleexist}  & \text{\moduleexist} \\
	\hline
	
	seglearn \pythonprimary &  \text{}  &  \text{}   &   \text{}   &   \text{}   &   \text{}  &    \text{}  &  \text{}   
	&  \text{}   &  \text{}   &  \text{} &  \text{\moduleexist}  &  \text{} 
	&  \text{}  &  \text{\moduleexist} 
	&  \text{}   &  \text{}  & \text{} \\
	\hline
	
	Shogun \cppprimary &  \text{\moduleexist}  &  \text{\moduleexist}   &   \text{\moduleexist}   &   \text{}   &   \text{}  &    \text{}  &  \text{}   
	&  \text{}   &  \text{}   &  \text{} &  \text{}  &  \text{} 
	&  \text{}  &  \text{\moduleexist} 
	&  \text{}   &  \text{\moduleexist}  & \text{} \\
	\hline
	
	sktime  \pythonprimary &  \text{\moduleexist}  &  \text{\moduleexist}   &   \text{}   &   \text{\moduleexist}   &   \text{\moduleexist}  &    \text{}  &  \text{}   
	&  \text{}   &  \text{\moduleexist}   &  \text{\moduleexist} &  \text{\moduleexist}  &  \text{\moduleexist} 
	&  \text{}  &  \text{\moduleexist} 
	&  \text{\moduleexist}   &  \text{\moduleexist}  & \text{\moduleexist} \\
	\hline
	
	sktime-dl \pythonprimary &  \text{\moduleexist}  &  \text{\moduleexist}   &   \text{}   &   \text{}   &   \text{}  &    \text{}  &  \text{}   
	&  \text{}   &  \text{}   &  \text{} &  \text{}  &  \text{} 
	&  \text{}  &  \text{} 
	&  \text{\moduleexist}   &  \text{}  & \text{} \\
	\hline
	
	statsmodels \pythonprimary &  \text{\moduleexist}  &  \text{}   &   \text{}   &   \text{}   &   \text{}  &    \text{}  &  \text{}   
	&  \text{\moduleexist}   &  \text{\moduleexist}   &  \text{\moduleexist} &  \text{}  &  \text{} 
	&  \text{\moduleexist}  &  \text{\moduleexist} 
	&  \text{\moduleexist}   &  \text{\moduleexist}  & \text{\moduleexist} \\
	\hline
	
	
	
	stumpy \pythonprimary &  \text{}  &  \text{}   &   \text{}   &   \text{}   &   \text{\moduleexist}  &    \text{\moduleexist}  &  \text{}   
	&  \text{}   &  \text{}   &  \text{} &  \text{}  &  \text{} 
	&  \text{}  &  \text{} 
	&  \text{\moduleexist}   &  \text{}  & \text{} \\
	\hline

	SUOD \pythonprimary &  \text{}  &  \text{}   &   \text{}   &   \text{\moduleexist}   &   \text{}  &    \text{}  &  \text{}   
	&  \text{}   &  \text{}   &  \text{} &  \text{}  &  \text{} 
	&  \text{\moduleexist}  &  \text{} 
	&  \text{\moduleexist}   &  \text{\moduleexist}  & \text{\moduleexist} \\
	\hline

	Telemanom \pythonprimary &  \text{}  &  \text{}   &   \text{}   &   \text{\moduleexist}   &   \text{}  &    \text{}  &  \text{}   
	&  \text{}   &  \text{}   &  \text{} &  \text{}  &  \text{} 
	&  \text{}  &  \text{\moduleexist} 
	&  \text{}   &  \text{\moduleexist}  & \text{\moduleexist} \\
	\hline
	
    TODS  \pythonprimary  &  \text{}  &  \text{}   &   \text{}   &   \text{\moduleexist}   &   \text{}  &    \text{\moduleexist}  &  \text{}   
	&  \text{\moduleexist}   &  \text{}   &  \text{\moduleexist} &  \text{}  &  \text{} 
	&  \text{}  &  \text{\moduleexist} 
	&  \text{\moduleexist}   &  \text{\moduleexist}  & \text{} \\
	\hline
	
	tftb \pythonprimary &  \text{}  &  \text{}   &   \text{}   &   \text{}   &   \text{}  &    \text{}  &  \text{}   
	&  \text{}   &  \text{}   &  \text{\moduleexist} &  \text{\moduleexist}  &  \text{} 
	&  \text{\moduleexist}  &  \text{} 
	&  \text{}   &  \text{\moduleexist}  & \text{\moduleexist} \\
	\hline
	
	tsfresh \pythonprimary  &  \text{}  &  \text{}   &   \text{}   &   \text{}   &   \text{}  &    \text{}  &  \text{}   
	&  \text{}   &  \text{\moduleexist}   &  \text{} &  \text{\moduleexist}  &  \text{} 
	&  \text{}  &  \text{} 
	&  \text{\moduleexist}   &  \text{\moduleexist}  & \text{} \\
	\hline
	
	tslearn \pythonprimary  &  \text{}  &  \text{\moduleexist}   &   \text{\moduleexist}   &   \text{}   &   \text{}  &    \text{\moduleexist}  &  \text{}   
	&  \text{}   &  \text{}   &  \text{} &  \text{\moduleexist}  &  \text{\moduleexist} 
	&  \text{\moduleexist}  &  \text{\moduleexist} 
	&  \text{}   &  \text{\moduleexist}  & \text{} \\
	\hline
	
	tsml 	\juliaprimary &  \text{\moduleexist}  &  \text{\moduleexist}   &   \text{}   &   \text{}   &   \text{}  &    \text{}  &  \text{}   
	&  \text{}   &  \text{\moduleexist}   &  \text{} &  \text{}  &  \text{} 
	&  \text{}  &  \text{\moduleexist} 
	&  \text{\moduleexist}   &  \text{\moduleexist}  & \text{} \\
	\hline

	
	Weka \javaprimary &  \text{\moduleexist}  &  \text{\moduleexist}   &   \text{\moduleexist}   &   \text{\moduleexist}   &   \text{}  &    \text{}  &  \text{}   
	&  \text{}   &  \text{\moduleexist}   &  \text{\moduleexist} &  \text{}  &  \text{} 
	&  \text{\moduleexist}  &  \text{\moduleexist} 
	&  \text{\moduleexist}   &  \text{\moduleexist}  & \text{\moduleexist} \\
	\hline

    \centering \bf{Total (60)} &  \text{32}  &  \text{14}   &   \text{13}   &   \text{21}   &   \text{5}  &    \text{8}  &  \text{8}   
	&  \text{7}   &  \text{23}   &  \text{24} &  \text{24}  &  \text{9} 
	&  \text{22}  &  \text{31} 
	&  \text{27}   &  \text{41}  & \text{39} \\
	\hline
	

\end{longtable}

\end{landscape}

\end{document}